# Measuring the originality of intellectual property assets based on machine learning outputs

**Author:** Sébastien Ragot

**Professional address:** E. Blum & Co. Ltd., Patent and Trademark Attorneys VSP, Vorderberg 11, 8044 Zürich, Switzerland.

**Abstract**

Originality criteria are frequently used to compare assets and, in particular, to assess the validity of intellectual property (IP) rights such as copyright and design rights. In this work, the originality of an asset is formulated as a function of the distances between this asset and its comparands, using concepts of maximum entropy and surprisal analysis. Namely, the originality function is defined according to the surprisal associated with a given asset. Creative assets can be justifiably compared to particles that repel each other via an electrostatic-like pair potential. This allows a very simple, suitably bounded formula to be obtained, in which the originality of an asset writes as the ratio of a reference energy to an interaction energy imparted to that asset. In particular, the originality of an asset can be expressed as a ratio of two average distances, i.e., the harmonic mean of the distances from this asset to its comparands divided by the harmonic mean of the distances between the sole comparands. Accordingly, the originality of objects such as IP assets can be simply estimated based on distances computed thanks to unsupervised machine learning techniques or other distance computation algorithms. Application is made to various types of assets, including emojis, typeface designs, paintings, and novel titles.

## 1. Introduction

The concept of originality plays an important role in human activities. People naturally tend to notice originality, whether in arts, at work, or in stores. In turn, the perceived originality of a thing (an object, an item, or a concept) often impacts its fate, i.e., how people react to it, whether they approve it, buy it, etc. [1]. In particular, originality can act as an amplifier, e.g., intensifying word-of-mouth phenomena, in either a positive or negative way [2].

By definition, the perceived originality is subjective: the human brain is naturally inclined to recognize originality, yet without measuring it objectively. Moreover, human capacities for perceiving originality are limited; they fade as the number of compared objects increases. Therefore, it is sometimes necessary to establish rules to assess originality. This is notably true in intellectual property (IP) as originality is a criterion that is used in the appreciation of the validity of copyright, industrial design rights, and related IP rights, even though the concept of originality is not universally nor unambiguously defined in the IP arena [3].

More generally, measuring originality is considered a challenging problem, regardless of the field of study [4, 5, 6, 7, 8, 9, 10, 11]. A major difficulty already arises in the definition of originality in the specific context of the study. General dictionaries define originality as the quality of being novel or unusual, i.e., the quality of being special and not the same as anything else. Based on this definition, one understands that an object can be considered original if it noticeably departs from comparable objects. Thus, in general, the originality of a thing may conceivably be regarded as a function of the distances between this thing and other comparable things.

Now, machine learning (ML) algorithms can be used to compute distances between assets, including IP assets such as works and designs. The more important the differences (or dissimilarities) between two assets, the larger the distances between them. Accordingly, one may want to use distances as computed by ML algorithms to evaluate the originality of objects. However, comparing such distances becomes rapidly intractable as the number of assets considered increases. Therefore, analytics are needed. Ideally, a simple formula would be desired to measure the originality of each IP asset.

This is precisely the purpose of the paper, which proposes a simple statistical measure of the degree of originality of assets, based on outputs from feature extraction or distance computation algorithms. The originality function is devised based on concepts of maximum entropy and surprisal analysis. The resulting function can be applied to any type of digital content, provided that distances between the corresponding objects can be evaluated.

The paper is structured as follows. Section 2 describes the problem at hand and the method used to construct the originality function. Examples of applications are discussed in Sect. 3, which notably concern pictures representing emojis (Sect. 3.1), typeface designs (i.e., fonts, Sect. 3.2), paintings (Sect. 3.3), and novel titles (Sect. 3.4). Limitations and possible generalizations of the model are addressed in Sect. 4. Related works are briefly commented on in Sect. 5. The last section closes this paper with some concluding remarks.

## 2. Proposed model of originality function

### 2.1. Problem at hand

The aim is to achieve a mathematical definition of the originality, based on distances computed using ML-related techniques or related algorithms. Feature extraction algorithms allow input data to be converted into vectors. Various extraction algorithms exist, which allow a variety of digital content to be processed, such as images, 3D data, sounds, and text documents, including product datasheets and patents [12, 13, 14]. Next, distances can be simply computed between points mapped by the vectors extracted from input datafiles. Accordingly, one may leverage ML-related techniques to estimate distances between any kind of IP assets, for example datafiles corresponding to copyrighted works (images, text, etc.), trademarks (including figurative marks, wordmarks, and sound trademarks), and design or utility patents. Besides ML-related techniques, other algorithms allow distances to be directly computed between input data, such as the so-called Edit distance for strings.

However, the analysis of such distances becomes rapidly complicated as the number of compared items increases. Comparing $N$ items gives rise to $N(N-1)/2$ pair distances. In particular, assessing a given asset with respect to $N – 1$ comparands gives rise to $N – 1$ distances, which must themselves be compared to the $(N-1)(N-2)/2$ distances involved between the sole comparands to arrive at meaningful conclusions. For example, assessing a given asset in view of only 10 comparands gives rise to 10 distances, which must be compared with 45 reference distances. Now, in practice, hundreds, if not thousands or more, of comparands may possibly be involved. Therefore, there is a need for a simple statistical measure of the originality of IP assets.

Having such a statistical measure may notably be useful to IP offices, IP professionals, and, more generally, to all IP players willing to assess the originality of IP assets such as copyrighted works and design rights. IP professionals may for instance need to assess the originality of given assets in view of prior art, i.e., in a time-ordered fashion. Besides, a statistical measure of the originality would be useful to anyone wishing to compare originalities of assets, regardless of their creation dates (or otherwise applicable dates), i.e., in a time-agnostic manner.

The problem at hand is now discussed in reference to Table 1, which shows distances obtained between selected emojis, in fact smiley faces in this example. Emojis are digital files of pictures (e.g., ideograms) that may potentially be protected by copyright, design rights, and/or trademarks. In this example, an unsupervised ML approach is used to compute distances between the selected smiley faces [15], using a feature extraction algorithm set to extract semantic features of the corresponding images [16]. The feature extraction algorithm converts each smiley into a respective semantic vector. Then, Euclidean distances are computed between points mapped by such vectors. The distances shown in Table 1 are normalized: the initial distances are divided by their arithmetic mean, such that the mean distance in the table is 1.

|   | 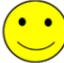 | 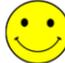 | 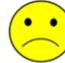 | 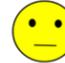 | 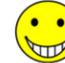 | 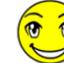 |
|---|---|---|---|---|---|---|
| 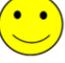 | 0. | 0.5 | 0.7 | 0.8 | 1. | 1.2 |
| 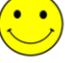 | 0.5 | 0. | 0.8 | 0.9 | 1. | 1.2 |
| 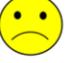 | 0.7 | 0.8 | 0. | 0.8 | 1.1 | 1.3 |
| 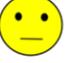 | 0.8 | 0.9 | 0.8 | 0. | 1.1 | 1.3 |
| 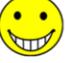 | 1. | 1. | 1.1 | 1.1 | 0. | 1.3 |
| 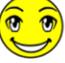 | 1.2 | 1.2 | 1.3 | 1.3 | 1.3 | 0. |

Table 1 (single-column picture, color): normalized distances between emojis computed from vectors obtained via a feature extraction algorithm set to extract semantic features of the corresponding images [16].

The resulting distances can be quite easily compared in this example, because of the small number of emojis considered. The 6 items considered result in 15 unique pair distances, which makes it relatively easy to estimate the relative originality of each emoji: the larger the distance to other emojis, the more original it is, *a priori*. What is more difficult, however, is to quantify the originality of each item. All the more, doing this becomes rapidly complicated as the number of selected IP assets increase. Now, when assessing the originality of IP assets such as pictorial representations, hundreds or thousands of potentially relevant representations may already exist, which may have to be considered. This requires a systematic and computerized approach, hence the need of a simple originality measure.

### 2.2. Proposed originality function

Intuitively, one understands that the originality of an asset may be written as the ratio of two average distances: the average distance between that asset and its comparands divided by the average distance between the comparands. This, however, immediately raises questions. E.g., which average formula (arithmetic, geometric, etc.) should we use? How can we enforce correct bounds of the resulting function? For instance, one would expect the originality score to be zero when an asset collides with another, e.g., when it is identical to a prior asset.

The approach proposed herein to address such questions is inspired by the Boltzmann distribution in statistical mechanics; it relies on concepts of information theory and surprisal analysis. Namely, assets are compared to particles that electrostatically repel each other, like same-sign point charges. The underlying idea is that human creativity would make it surprising to find two independently created assets to be very close to each other, also in the space spanned by outputs of a feature extraction algorithm.

Consider a set of $N$ assets of interest, which can be associated with a set of extracted points $\{\mathbf{r}_i | i \in \mathbb{N}, 1 \leq i \leq N\}$, hereafter noted $\{\mathbf{r}_i\}$ for simplicity. Such assets are assumed to be subject to repulsive pair interactions, such that the total interaction energy $U(\mathbf{r}_1, \ldots, \mathbf{r}_N)$ of the assets can be written as

$$U(\mathbf{r}_1, \ldots, \mathbf{r}_N) = \frac{1}{2} \sum_{i=1}^{N} \sum_{\substack{j=1 \\ j \neq i}}^{N} u_{ij}, \tag{1}$$

where $u_{ij} = u(|\mathbf{r}_j - \mathbf{r}_i|) = u(r_{ij})$ denotes a repulsive interaction between a given pair $\{i, j\}$ of assets. Each term $u_{ij}$ is positive, such that the total interaction energy $U$ is positive too. The pair interaction potential must tend to infinity when $\mathbf{r}_j \to \mathbf{r}_i$ (the two assets are identical at the limit $\mathbf{r}_j = \mathbf{r}_i$) since identical assets give rise to a same vector. Conversely, the pair potential must tend to zero when the distance between $\mathbf{r}_j$ and $\mathbf{r}_i$ becomes infinite. That is,

$$\lim_{r \to \infty} u(r) = 0, \text{ and}$$
$$\lim_{r \to 0} u(r) = \infty. \tag{2}$$

A suitable pair potential is derived in the next section.

The total interaction energy $U$ can be partitioned into single-asset contributions. A simple partition is the following:

$$U(\mathbf{r}_1, \ldots, \mathbf{r}_N) = \sum_{i}^{N} \left\{ \frac{1}{2} \sum_{\substack{j=1 \\ j \neq i}}^{N} u_{ij} \right\} = \sum_{i=1}^{N} U_i. \tag{3}$$

That is, $U$ can be written as a sum of contributions $U_i = \frac{1}{2}\sum_{j=1 \, \{j \neq i\}}^{N} u_{ij}$, each involving interactions between a given asset $i$ and the remaining assets.

At present, what is needed is the probability distribution of $U$, assuming that the total interaction energy has a positive and specified mean $\langle U \rangle > 0$. A convenient form is provided by the maximum entropy distribution $p(U)$, which, under the constraints that $\int_U p(U) dU = 1$ and $\int_U U p(U) dU = \langle U \rangle$, is known to be:

$$p(U) = \frac{1}{\langle U \rangle} e^{-U/\langle U \rangle}. \tag{4}$$

As expected, the probability distribution $p(U)$ tends to zero when $U$ tends to infinity and is maximal at $U = 0$, in accordance with the assumption that creativity causes the created assets to depart from each other.

The probability $p(U(\mathbf{r}_1, \ldots, \mathbf{r}_N)) = e^{-U(\mathbf{r}_1, \ldots, \mathbf{r}_N)/\langle U \rangle}/\langle U \rangle$ represents the probability associated with a state where the assets are located in $\{\mathbf{r}_i\}$. Now, given repulsive pair interactions occurring between the assets, one may *a priori* expect to find such assets infinitely far from each other, in which case the total interaction energy (call it $U_\infty$) vanishes. The probability $p(U_\infty) = 1/\langle U \rangle$ can thus be associated with a state of the system in which the assets are infinitely far from each other.

Using the above notations, one may now define the surprisal $S(\{\mathbf{r}_i\}|\langle U \rangle)$ of finding the assets actually located at given positions $\{\mathbf{r}_i\}$ as a measurement of the deviation of $p(U(\mathbf{r}_1, \ldots, \mathbf{r}_N))$ from $p(U_\infty)$, as in surprisal analysis. Namely:

$$S(\{\mathbf{r}_i\}|\langle U \rangle) = -\ln \frac{p(U(\mathbf{r}_1, \ldots, \mathbf{r}_N))}{p(U_\infty)}. \tag{5}$$

Replacing now the probabilities involved in Eq. (5) by their expressions, one obtains:

$$S(\{\mathbf{r}_i\}|\langle U\rangle) = -\ln\exp\left(-\frac{U(\mathbf{r}_1,\ldots,\mathbf{r}_N)}{\langle U\rangle}\right) = \frac{U(\mathbf{r}_1,\ldots,\mathbf{r}_N)}{\langle U\rangle} \tag{6}$$

Thus, the surprisal $S(\{\mathbf{r}_i\}|\langle U\rangle)$ is simply the ratio of the total interaction energy $U(\mathbf{r}_1,\ldots,\mathbf{r}_N)$ to the average energy $\langle U\rangle$. The surprisal is positive, tends to zero when the assets are infinitely far from each other (since $U_\infty = 0$), and tends to infinity when any two assets collide, i.e., when any $u_{ij} \to \infty$, see Eqs. (1) and (2). In other words, given that creative assets repel each other and are therefore expected to be, *a priori*, infinitely far from each other, it would be infinitely surprising to find any asset colliding with any one of the other assets.

As defined above, the surprisal function $S(\{\mathbf{r}_i\}|\langle U\rangle)$ pertains to the whole set of assets. However, Eq. (3) can be used to partition the surprisal function into single-asset terms, noted $S(\mathbf{r}_k|\langle U\rangle)$, i.e.,

$$S(\mathbf{r}_k|\langle U\rangle) = \frac{U_k}{\langle U\rangle}. \tag{7}$$

The surprisal $S(\mathbf{r}_k|\langle U\rangle)$ associated with a single asset $k$ is positive, tends to zero when the asset $k$ is infinitely far from each of the other assets ($U_k = 0$ in that case) and tends to infinity when the asset $k$ collides with any of the other assets.

Now, one expects the originality function associated with a given asset $k$ to behave in the exact opposite manner. Thus, one may define the originality $\mathcal{O}(\mathbf{r}_k|\langle U\rangle)$ of a particular asset $k$ as the reciprocal of the surprisal $S(\mathbf{r}_k|\langle U\rangle)$:

$$\mathcal{O}(\mathbf{r}_k|\langle U\rangle) = \frac{1}{S(\mathbf{r}_k|\langle U\rangle)} = \frac{\langle U\rangle}{U_k}. \tag{8}$$

As defined above, the originality $\mathcal{O}(\mathbf{r}_k|\langle U\rangle)$ of a particular asset $k$ is simply the ratio of $\langle U\rangle$ to $U_k$. This function is positive. Conveniently enough, it tends to zero when this asset $k$ collides with any of the other assets and tends to infinity when the asset $k$ is infinitely far from each of the other assets.

In Eq. (8), the originality function $\mathcal{O}(\mathbf{r}_k|\langle U\rangle)$ is defined with respect to the average energy $\langle U\rangle$, which can be regarded as a proportionality constant that impacts the value of $\mathcal{O}(\mathbf{r}_k|\langle U\rangle)$, without, however, affecting its behavior. Thus, any reference energy may potentially be used instead of $\langle U\rangle$. Now, it is often desired to evaluate the originality of an asset in view of its sole comparands, e.g., the prior assets when dealing with IP matters. To that aim, one may introduce

$$\mathcal{O}(\mathbf{r}_k|U_{\text{ref},k}) = \frac{U_{\text{ref},k}}{U_k}, \tag{9}$$

where $U_{\text{ref},k}$ is a positive, reference energy, pertaining to selected comparands of the asset $k$, e.g., the remaining assets located in $\{\mathbf{r}_1, \ldots \mathbf{r}_{k-1}, \mathbf{r}_{k+1}, \ldots, \mathbf{r}_N\}$. That is, the reference energy $U_{\text{ref},k}$ may be defined similar to $U(\mathbf{r}_1, \ldots, \mathbf{r}_N)$ in Eq. (3), yet excluding any pair involving the asset $k$. Moreover, to ease the comparisons, quantitatively, this reference energy $U_{\text{ref},k}$ may be scaled to be commensurate with $U_k$. I.e., one may notably define $U_{\text{ref},k}$ as:

$$U_{\text{ref},k} = \frac{1}{2(N-2)} \sum_{\substack{i=1 \\ i \neq k}}^{N} \sum_{\substack{j=1 \\ j \neq i,k}}^{N} u_{ij}, \tag{10}$$

where the factor $1/(N-2)$ compensates for the fact that $U_k$ involves $(N-1)$ pair interactions, while $U_{\text{ref},k}$ involves $(N-1)(N-2)$ pairs.

Doing so, the originality function $\mathcal{O}(\mathbf{r}_k | U_{\text{ref},k})$ rewrites as:

$$\mathcal{O}(\mathbf{r}_k | U_{\text{ref},k}) \equiv \mathcal{O}_k = \frac{1}{(N-2)} \frac{\sum_{i=1\{i \neq k\}}^{N} \sum_{j=1\{j \neq i,k\}}^{N} u_{ij}}{\sum_{j=1\{j \neq k\}}^{N} u_{kj}}. \tag{11}$$

The originality $\mathcal{O}(\mathbf{r}_k | U_{\text{ref},k})$ of a single asset is hereafter noted $\mathcal{O}_k$, to ease the exposition. Owing to the normalization choice made in Eq. (11), the originality $\mathcal{O}_k$ is equal to 1 when the energy $U_k$ is equal to the reference energy $U_{\text{ref},k}$. Any value of $\mathcal{O}_k$ that is larger than 1 denotes an above-average originality (meaning that the corresponding energy $U_k$ is less than the mean interaction energy $U_{\text{ref},k}$ of the comparands), whereas $\mathcal{O}_k < 1$ denotes sub-original assets, as illustrated in FIG. 1.

### 2.3. Pair interaction energy

At present, a suitable expression of the pair interaction potential is derived. At least two comparands are needed to evaluate the originality of an asset; $N$ must be strictly larger than 2 in Eq. (11). Consider that only three assets are involved in total, for simplicity, and assume that the vector extracted for one of the three assets (say asset $k$) maps a point located exactly in the middle of the line segment of length $r_0$ extending between the two other extracted points. As the asset $k$ is located at a distance $r_0/2$ of each of the two other points (themselves at a distance $r_0$ from each other), one may want to set the originality score of the asset $k$ to $1/2$. Now, according to Eq. (11), the originality of the asset $k$ is $\mathcal{O}_k = 2\,u(r_0)/(2\,u(r_0/2))$. Imposing it to be equal to $1/2$ yields a functional equation $u(r_0) = u(r_0/2)/2$, the simplest solution of which is $u(r_0) = 1/r_0$ [17].

The resulting pair interaction potential is thus similar to the electrostatic repulsion between like charges, i.e.,

$$u(|\mathbf{r}_j - \mathbf{r}_i|) = \frac{1}{|\mathbf{r}_j - \mathbf{r}_i|} = \frac{1}{r_{ij}}, \tag{12}$$

and fulfills the constraints set in Eq. (2).

Substituting $u_{ij}$ with $1/r_{ij}$ in Eq. (11) yields the final expression of the originality function, i.e.,

$$\mathcal{O}_k = \frac{1}{(N-2)} \frac{\sum_{i=1\{i \neq k\}}^N \sum_{j=1\{j \neq i,k\}}^N \frac{1}{r_{ij}}}{\sum_{j=1\{j \neq k\}}^N \frac{1}{r_{kj}}}, \tag{13}$$

which will be used for practical calculations in Sect. 3.

### 2.4. Properties of the originality function

The originality measure proposed in Eqs. (9), (11), and (13) has the following properties.

Normalization. The reference energy term $U_{\text{ref},k}$, Eq. (10), can also be formulated as $U_{\text{ref},k} = (U - 2U_k)/(N - 2)$, such that Eq. (9) can be rewritten as

$$\mathcal{O}(\mathbf{r}_k | U_{\text{ref},k}) \equiv \mathcal{O}_k = \frac{(U - 2U_k)}{(N-2)U_k}, \tag{14}$$

where $U \equiv U(\mathbf{r}_1, \ldots, \mathbf{r}_N)$ is the total interaction energy. This implies that $(N - 2)\mathcal{O}_k + 2 = U/U_k$, whereby the terms $\mathcal{O}_k$ are subject to the following normalization condition:

$$\sum_{k=1}^N \frac{1}{(N-2)\mathcal{O}_k + 2} = 1. \tag{15}$$

Limits. As noted earlier, the originality value $\mathcal{O}_k$ obtained for an asset $k$ of interest (mapped to a vector $\mathbf{r}_k$) tends to zero when any of the distances $r_{kj}$ between the asset $k$ in $\mathbf{r}_k$ and its comparands in $\{\mathbf{r}_j\}$ tends to zero. It can be set to zero at the limit $r_{kj} = 0$, which denotes a collision between the assets $k$ and $j$. Note, an originality value of zero denotes a lack of novelty in view of at least one of the comparands $\{j\}$ when the latter are prior assets. Conversely, the farther the asset $k$ from the remaining assets, the lower its energy component $U_k$, and the larger its originality value. The originality of an asset that is infinitely far from each of the other assets is infinite, see Eqs. (9), (11), and (13).

Impact of comparands. The originality score obviously depends on the available or selected comparands. However, the originality value $\mathcal{O}_k$ of an asset $k$ is essentially impacted by its closest comparands. Farther neighbors have a progressively decreasing impact on the originality value of item $k$. However, remote comparands still impacts the reference energy $U_{\text{ref},k}$ and thus the originality values.

In this regards, Eq. (13) can be regarded as a scaled ratio of two harmonic means. As such, one expects it to be quite sensitive to minima of each of the numerator and the denominator. In particular, the originality value $\mathcal{O}_k$ may become infinitely small if the asset of interest happens to be infinitely close to any of the comparands, which is a desired property, as noted above. Yet, the originality value $\mathcal{O}_k$ may also become infinitely large if any two comparands come infinitesimally close to each other, as this proximity impacts the reference energy $U_{\text{ref},k}$.

Thus, minimal care should be taken in selecting the comparands, be it to prevent doubloons. In this work, the assets considered are assumed to be unique and therefore give rise to distinct extracted points $\{\mathbf{r}_j\}$. Any comparand relevant to the investigated asset $k$ should be included, in principle. However, in practice, because only a limited selection of objects can be compared, the originality values that result are always relative (i.e., relative to the set of comparands considered). Moreover, originality values computed for selected objects must further be compared to each other for them to make sense. Additional remarks are compounded in Sect. 4.

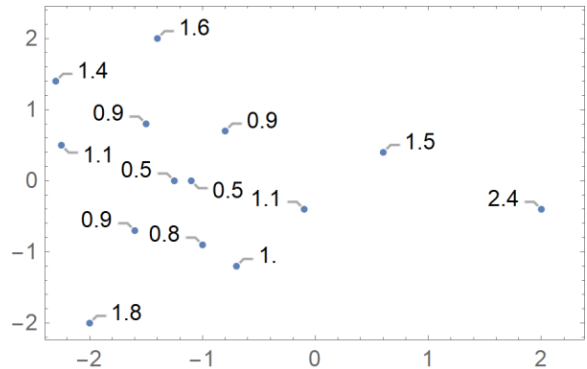

FIG. 1 (Single-column, color): 2D plot of values of the originality function $\mathcal{O}_k$, Eq. (13), evaluated for arbitrary points in a 2D space. The originality scores of the corresponding items (i.e., mapped to the respective points) is evaluated for each item with respect to all of the remaining items.

Figure 1 shows values taken by the originality function $\mathcal{O}$ as defined in Eq. (13) for an example set $\mathcal{S}$ of points distributed in a 2D space. All points are assumed to lie in a 2D vector space in this example, for simplicity. The originality values $\mathcal{O}_k$ are computed by successively comparing each point $\mathbf{x}_k$ of the set $\mathcal{S}$ to the remaining points $\{\mathbf{x}_1, \ldots, \mathbf{x}_{k-1}, \mathbf{x}_{k+1}, \ldots, \mathbf{x}_n\}$. Unsurprisingly, remote points have larger originality values, while points in the denser region of the plot give rise to lower originality scores.

This is consistent with the fact that most similar items give rise to vectors mapped to close points, whereas an original item corresponds to a point that is farther from the remaining points.

## 3. Examples of applications of the originality function

Various examples of application are now discussed. These examples relate to emojis, typeface designs (fonts), and paintings. In each case, vectors obtained from a feature extraction algorithm are used to compute Euclidean distances, based on which the originality is evaluated for each asset, according to Eq. (13). The feature extraction algorithm is set to extract semantic features of the considered images; similar results are obtained when considering an extraction based on pixel values (not reported). Additional examples are discussed *in fine*, which concern textual content and therefore require other extraction schemes or distance computation algorithms.

### 3.1. Emojis

Following up on issues noted earlier in reference to Table 1, originality values of the emojis are now computed using Eq. (13). The results are depicted as a bar chart in FIG. 2.

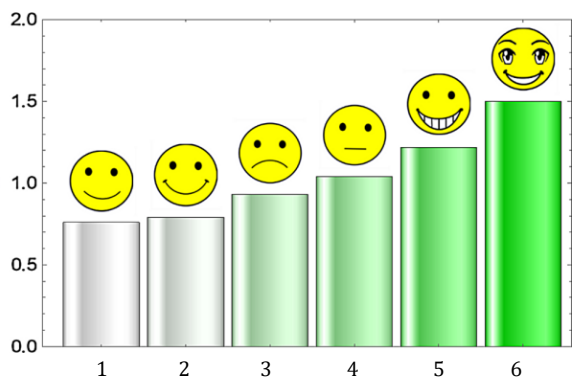

FIG. 2 (single-column, color): Bar chart of originality values computed from Eq. (13) for the emojis of Table 1. The originality values are evaluated for each item in view of all of the other items, in a time-agnostic manner.

The bar chart shown in FIG. 2 reflects originality values computed for each emoji in view of all the others, in a time-agnostic manner, i.e., irrespective of potential creation dates. In variants, originality values may also be evaluated in a time-ordered fashion, taking into account the dates applicable to each items, as IP players may want to do to estimate the originality of IP assets. The originality scores of the items are ordered in ascending order in FIG. 2. The scores are impacted by the relative rarity and size of the features shown in the pictures. E.g., the arc symbolizing a smile is relatively common; it appears in the first and second items, and also forms part of the open-mouth smiles seen in the fifth and sixth items. A flipped version of this arc is seen in the third item, making a frowning face. The horizontal line segment of the neutral face (fourth item) is rare but thin. On the contrary, the teeth and the "manga-like" eyes are not only rare but also relatively large; they occupy a substantial portion of the fifth and sixth items.

### 3.2. Application to typeface designs (fonts)

The terms "fonts" and "typeface designs" are often used interchangeably. In modern usage, "font" is typically used to refer to the digital file used by a computer system to display the corresponding glyphs. However, a font more specifically refers to a particular size, weight, and style of a typeface. For example, the Arial typeface comprises many styles, including Regular, Italic, and Bold [18].

Not all countries allow the artistic design of a typeface to be protected by copyright law. E.g., the law in Germany and the U.K. protects typeface designs, whereas copyright in the U.S. only protects the corresponding digital files. Besides copyright, protection for typeface designs may be obtained via design patents, whereas a wordmark only protects the name of a typeface [19].

In the following, the originality of selected typefaces is assessed using a simple pipeline. In total, 83 popular typefaces were considered, wherein only the regular font styles were included, it being noted that some of the regular font styles have an italic-like semblance.

First, a comparison basis is needed. To that aim, one may for example use a pangram— a sentence that involves all letters of the English alphabet, as routinely done to compare typefaces. Instances of the resulting text are displayed in Table 2 for a few popular typefaces.

| Selected typeface | Fixed–width pangram |
|---|---|
| Franklin Gothic Medium | The quick brown fox jumps over the lazy dog. |
| Impact | The quick brown fox jumps over the lazy dog. |
| Arial | The quick brown fox jumps over the lazy dog. |
| Lucida Console | The quick brown fox jumps over the lazy dog. |
| Comic Sans MS | The quick brown fox jumps over the lazy dog. |
| Calibri | The quick brown fox jumps over the lazy dog. |
| Corbel | The quick brown fox jumps over the lazy dog. |
| Candara | The quick brown fox jumps over the lazy dog. |

Table 2 (single-column picture, black and white): Examples of typefaces and corresponding pangrams.

Semantic features of images of all the pangrams are then extracted, prior to computing a distance matrix according to the extracted vectors. Originality values are finally computed thanks to Eq. (13). Table 3 aggregates originality scores obtained for all typefaces considered.

Clearly, the rightmost typefaces in Table 3 look more original than the leftmost typefaces. In particular, Impact and Mistral are found among the most original typefaces, whereas Segoe UI, Gadugi, Arial, and Calibri are among the least original typefaces of the selection.

Of course, such results must be taken with a grain of salt, given the limited selection of typefaces retained. There are thousands of typefaces; an exhaustive study of the originality of typeface designs (which is not the purpose of this paper) should take as many fonts as possible into account. Thus, the results presented in Table 3 reflect relative originality scores for the sole typefaces selected. In particular, italic-like typefaces (like Freestyle Script or Mistral) obtain higher originality scores simply because they are rarer than plain-like typefaces in the selection considered.

| Typeface | Score | Typeface | Score | Typeface | Score | Typeface | Score |
|---|---|---|---|---|---|---|---|
| Gadugi | 0.65 | IBM Plex Serif | 0.84 | Mongolian Baiti | 1. | Oswald | 1.16 |
| Segoe UI | 0.66 | Microsoft Sans Serif | 0.85 | PMingLiU–ExtB | 1.01 | Cousine | 1.17 |
| Malgun Gothic | 0.68 | Sitka Heading | 0.86 | Segoe Print | 1.01 | Playfair Display | 1.21 |
| Microsoft PhagsPa | 0.72 | Sitka Display | 0.87 | Source Code Pro | 1.01 | MV Boli | 1.27 |
| Clear Sans | 0.72 | Times New Roman | 0.87 | Tempus Sans ITC | 1.02 | Felipa | 1.28 |
| Corbel | 0.76 | Candara | 0.89 | Constantia | 1.02 | Kalam | 1.29 |
| Yu Gothic | 0.77 | Century Gothic | 0.89 | Sitka Text | 1.03 | SimSun | 1.35 |
| Lato | 0.77 | Trebuchet MS | 0.89 | Inconsolata | 1.03 | Pristina | 1.43 |
| Roboto | 0.78 | Source Serif Pro | 0.89 | Lucida Handwriting | 1.04 | Garamond | 1.44 |
| Source Sans Pro | 0.78 | Lucida Console | 0.91 | Book Antiqua | 1.04 | Freestyle Script | 1.47 |
| IBM Plex Sans | 0.78 | Titillium Web | 0.92 | Yanone Kaffeesatz | 1.05 | Ink Free | 1.55 |
| Dubai | 0.79 | Sitka Banner | 0.92 | Microsoft Yi Baiti | 1.06 | Bradley Hand ITC | 1.57 |
| Cambria | 0.8 | Microsoft Himalaya | 0.94 | Georgia | 1.06 | Kristen ITC | 1.58 |
| Microsoft JhengHei | 0.81 | MingLiU–ExtB | 0.94 | Gabriola | 1.07 | SimSun–ExtB | 1.76 |
| Calibri | 0.81 | Monotype Corsiva | 0.94 | Consolas | 1.08 | Alegreya SC | 1.87 |
| Bahnschrift | 0.83 | Droid Serif | 0.94 | Palatino Linotype | 1.13 | Juice ITC | 1.88 |
| Lucida Sans Unicode | 0.83 | MS Gothic | 0.96 | Economica | 1.13 | Impact | 1.97 |
| Verdana | 0.83 | Sylfaen | 0.96 | IBM Plex Mono | 1.14 | League Gothic | 2.04 |
| Century | 0.84 | Javanese Text | 0.96 | French Script MT | 1.14 | Papyrus | 2.05 |
| Arial | 0.84 | Bookman Old Style | 0.97 | Comic Sans MS | 1.15 | Mistral | 2.47 |
| Tahoma | 0.84 | Gentium Basic | 0.99 | Segoe Script | 1.15 | | |

Table 3 (double-column picture, black and white): Originality scores of an extended set of typefaces.

### 3.3. Paintings by Johannes Vermeer

At present, originality scores are computed for paintings by Johannes Vermeer [20] using Eq. (13), based on semantic features extracted from images of the paintings. The resulting scores are compounded in Table 4. The snippet images shown in Table 4 are cropped for layout purposes. It must be emphasized that the originality scores were computed based on distances between Vermeer's paintings only, irrespective of other artists' works. Thus, the resulting scores solely reflect the intrinsic originality of Vermeer's paintings.

Interestingly, the results obtained show that the most iconic paintings of the artist (e.g., "The Milkmaid", "Girl with a Pearl Earring", "View of Delft", and "The Lacemaker") tend to have larger originality scores. However, this finding cannot be generalized. For example, "Mona Lisa" is found to be the least original work, when applying the same procedure to paintings by Leonardo Da Vinci (not shown).

| Painting (snippet) | Title | Score | Painting (snippet) | Title | Score |
|---|---|---|---|---|---|
| 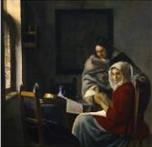 | *Girl Interrupted at Her Music* | 0.85 | 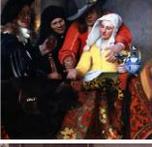 | *Officer with a Laughing Girl* | 0.97 |
| 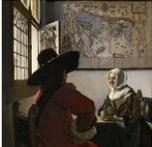 | *The Concert* | 0.89 | 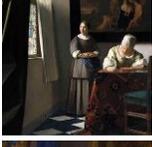 | *Portrait of a Young Woman* | 0.98 |
| 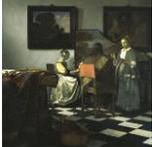 | *Woman Holding a Balance* | 0.89 | 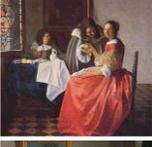 | *Girl Reading a Letter by an Open Window* | 0.99 |
| 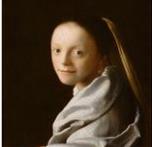 | *Christ in the House of Martha and Mary* | 0.89 | 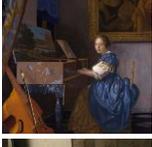 | *The Procuress* | 0.99 |
| 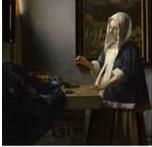 | *Lady Writing a Letter with her Maid* | 0.90 | 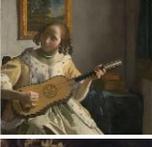 | *The Girl with the Wineglass* | 1.00 |
| 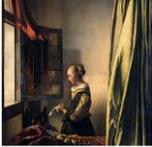 | *Lady Seated at a Virginal* | 0.91 | 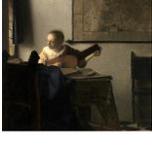 | *The Guitar Player* | 1.06 |
| 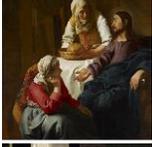 | *Woman with a Lute* | 0.92 | 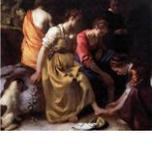 | *Diana and her Nymphs* | 1.06 |
| 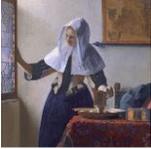 | *Woman with a Water Jug* | 0.92 | 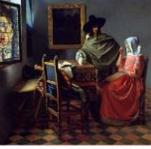 | *Mistress and Maid* | 1.07 |
| 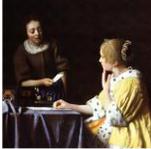 | *Allegory of the Catholic Faith* | 0.93 | 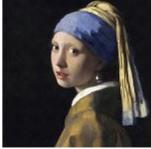 | *A Girl Asleep* | 1.07 |
| 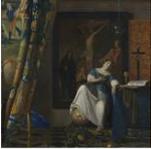 | *The Art of Painting* | 0.93 | 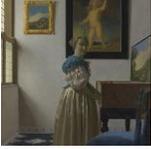 | *The Music Lesson* | 1.08 |
| 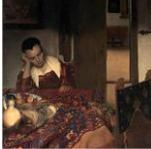 | *The Geographer* | 0.94 | 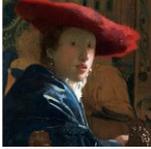 | *The Milkmaid* | 1.09 |
| 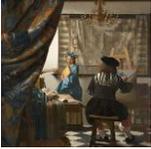 | *Woman with a Pearl Necklace* | 0.95 | 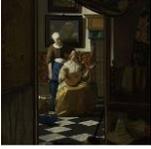 | *Woman in Blue Reading a Letter* | 1.09 |
| 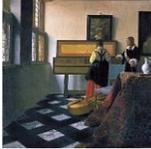 | *The Wine Glass* | 0.95 | 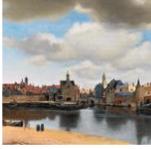 | *Girl with a Pearl Earring* | 1.12 |
| 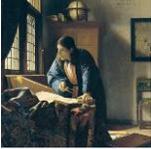 | *Lady Standing at a Virginal* | 0.97 | 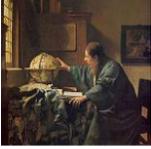 | *Girl with a Red Hat* | 1.16 |
| 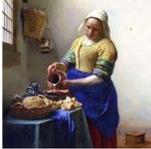 | *The Loveletter* | 0.97 | 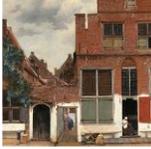 | *View of Delft* | 1.25 |
| 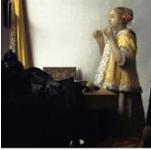 | *The Astronomer* | 0.97 | 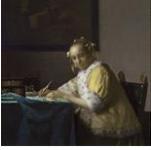 | *The Little Street* | 1.28 |
| 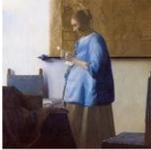 | *A Lady Writing a Letter* | 0.97 | 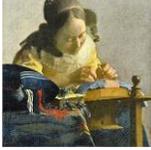 | *The Lacemaker* | 1.36 |

Table 4 (double-column picture, color): Originality scores obtained for paintings by Johannes Vermeer.

### 3.4. Textual content

Besides feature extraction algorithms for images, many other extraction algorithms exist. For instance, textual data can be compared semantically and/or syntactically. The very simple example in Table 5 below shows relative originality scores of four documents, each consisting of a few words only, for simplicity. The originality scores are computed based on vectors extracted using a word embedding scheme. I.e., words are transformed into vectors that capture the meaning of the words: the closer the vectors, the more similar the meanings. In this example, the documents relate to cats and dogs, subject to one exception ("*Jazz is music*"), which is accordingly found to be the most original document in this set.

| Textual content | Originality score |
|---|---|
| *The cat is black* | 0.74 |
| *The dog is white* | 0.74 |
| *A dog is not a cat* | 0.91 |
| *Jazz is music* | 1.04 |

Table 5 (single-column picture, black-and-white): Originality scores of a small set of textual documents, based on vectors extracted using a word embedding scheme.

However, an approach based on a word embedding scheme requires input words that are known to the embedding algorithm, for them to be suitably translated into vectors. Else, the algorithm typically converts any previously unseen word into the zero vector. Thus, word embedding schemes cannot be directly applied to assets such as brand names or novel titles, as the latter often involve made-up names and unconventional characters.

In such a situation, one may nevertheless resort to other extraction schemes, e.g., relying on a segmentation of the input text into characters or words, or on the edit distance (i.e., the Levenshtein distance), as made in Table 6 for English titles of selected novels of the twentieth century [21].

In detail, the first two columns of originality scores in Table 6 are obtained according to distances computed from vectors formed by the word frequencies and character frequencies (irrespective of the capitalization), as obtained in the bases formed by the union of the unique words and characters of the titles, respectively. The results in the last column are obtained from the Levenshtein distances between the titles. All originality scores are computed using Eq. (13).

The originality scores obtained by segmentation (first and second numerical columns in Table 6) are highly correlated: the Pearson's correlation coefficient and Spearman's rank correlation coefficient are equal to 0.89 and 0.88, respectively. The underlying extractions schemes tend to favor short titles as the latter necessarily result in higher character and word frequencies. Very different results are obtained when using the Levenshtein distance, which is impacted by the character counts and therefore favors long titles. The ranks of values in the last column are roughly anticorrelated with respect to values in the second numerical column (character frequency): the Spearman's rank correlation coefficient is -0.76.

This illustrates the fact that the originality values depend on the distance computation algorithm chosen. The originality values produced by the function defined by Eq. (13) remain consistent as long as the underlying extraction schemes are; dissimilar distance computation algorithms logically yield dissimilar originality values. Of course, any suitable extraction scheme may be contemplated.

|  | Originality values based on: | | |
| --- | --- | --- | --- |
| *Novel Title* | Word Frequency | Character Frequency | Edit Distance |
| *Harry Potter and the Order of the Phoenix* | 0.75 | 0.76 | 1.28 |
| *Harry Potter and the Goblet of Fire* | 0.78 | 0.76 | 1.09 |
| *Harry Potter and the Chamber of Secrets* | 0.78 | 0.78 | 1.18 |
| *Harry Potter and the Half - Blood Prince* | 0.83 | 0.74 | 1.29 |
| *Harry Potter and the Philosopher ' s Stone* | 0.83 | 0.79 | 1.32 |
| *Harry Potter and the Prisoner of Azkaban* | 0.78 | 0.86 | 1.24 |
| *Harry Potter and the Deathly Hallows* | 0.84 | 0.87 | 1.21 |
| *The Lion, the Witch and the Wardrobe* | 0.92 | 0.81 | 1.38 |
| *The Common Sense Book of Baby and Child Care* | 0.86 | 0.94 | 1.76 |
| *The Bridges of Madison County* | 0.95 | 0.91 | 1.17 |
| *The Name of the Rose* | 0.95 | 0.95 | 0.83 |
| *The Catcher in the Rye* | 0.99 | 1.00 | 0.87 |
| *One Hundred Years of Solitude* | 1.07 | 0.95 | 1.20 |
| *The Da Vinci Code* | 1.04 | 1.14 | 0.83 |
| *And Then There Were None* | 1.07 | 1.13 | 1.02 |
| *The Eagle Has Landed* | 1.04 | 1.17 | 0.89 |
| *You Can Heal Your Life* | 1.11 | 1.11 | 1.03 |
| *Charlotte 's Web* | 1.28 | 0.97 | 0.89 |
| *The Hite Report* | 1.11 | 1.14 | 0.76 |
| *The Little Prince* | 1.10 | 1.17 | 0.79 |
| *The Ginger Man* | 1.11 | 1.18 | 0.77 |
| *Anne of Green Gables* | 1.13 | 1.22 | 0.99 |
| *The Alchemist* | 1.23 | 1.21 | 0.77 |
| *Watership Down* | 1.46 | 1.15 | 0.93 |
| *The Hobbit* | 1.23 | 1.55 | 0.78 |
| *Lolita* | 1.90 | 2.18 | 0.92 |

Table 6 (double-column, black-and-white): Originality scores of selected novel titles, based on various extraction schemes (see text). Results are sorted according to the geometric mean of the frequency-based results (first two numerical columns).

## 4. Additional remarks: limitations, possible generalizations, and alternative definitions of originality

### 4.1. Limitations of the proposed definition of originality

As defined in Eq. (11) or (13), the originality value $\mathcal{O}_k$ of an asset $k$ of interest matches one's intuitive perception of originality, which is primarily determined by the distances (or dissimilarity) between this asset and the closest assets (e.g., the most resembling prior assets) among the comparands $\{\mathbf{r}_1, \ldots \mathbf{r}_{k-1}, \mathbf{r}_{k+1}, \ldots, \mathbf{r}_N\}$.

The analysis made in Sect. 2.3 advocates the use an $r^{-1}$ pair potential. However, the resulting originality function, Eq. (13), does not converge rapidly. Correlatively, the originality score obtained for a given asset is quite sensitive to the number of comparands considered. Still, any repulsive pair potential may be considered, in principle, provided that it meets the constraints set in Eq. (2). Thus, in variants to Eq. (12), one may prefer using a screened pair potential, e.g., $u(r) = \frac{1}{r}e^{-\alpha r}$ or any functional form that decreases faster than $r^{-1}$, e.g., $u(r) = r^{-n}, n \geq 2$. In practice, using a more rapidly converging pair potential increases the contrasts obtained between the originality values, compared with values obtained with Eq. (13).

Moreover, the originality function of Eq. (13) does not explicitly take into account combinations of features from the comparands. Such combinations, however, are indirectly accounted for in the extracted features, inasmuch as the semantic vector of an asset combining features of two other assets should be closer to these two assets than to less relevant assets.

Eq. (13) statistically accounts for multiple interactions; it reflects the originality of one asset interacting with many others. On the contrary, the originality of a design is often evaluated by comparing it to one prior design at a time; the question to be answered by the IP specialist is whether an asset of interest is sufficiently original with respect to each of the prior assets known, one at a time, though combinations of prior designs may sometimes be considered. To reflect this practice, one may modify Eq. (13) to limit both the numerator and the denominator to the $J$ closest comparand(s), where, e.g., $J = 1$ or 2. This way, an alternative definition of the originality function $\tilde{\mathcal{O}}_k$ is obtained, i.e.,

$$\tilde{\mathcal{O}}_k = \frac{1}{(N-1)} \frac{\sum_{i=1\{i \neq k\}}^{N} \sum_{j=1\{j \neq i,k\}}^{J} \frac{1}{\tilde{r}_{ij}}}{\sum_{j=1\{j \neq k\}}^{J} \frac{1}{\tilde{r}_{kj}}}, \tag{16}$$

where $\tilde{r}_{ij}$ is the $j$th order statistic of the set $\{r_{ij}\}$, that is, $\tilde{r}_{i1} < \tilde{r}_{i2} < \cdots < \tilde{r}_{iJ}$.

Beyond the issues noted above, it should be made clear that the originality functions proposed in this work cannot be used as a substitute for the judgment of an IP professional, especially as the definition of originality is not unambiguously defined in IP law. Plus, different jurisdictions may impose different originality tests. Thus, the originality functions proposed herein should rather be regarded as providing additional facts in the form of statistical measurements.

### 4.2. Possible generalization and alternative definitions

As said, Eq. (13) can be regarded as a scaled ratio of the harmonic mean of the distances between a given asset $k$ and the other assets to the harmonic mean of the distances between the remaining assets. Based on this observation, one may want to generalize this result to a scaled ratio of average distances, using any suitable definition of average distances.

The average distance of a given set of positive real numbers $\{d_1, \ldots, d_n\}$ can be computed thanks to the generalized mean formula, that is,

$$M_p(d_1, \ldots, d_n) = \left( \frac{1}{n} \sum_{i=1}^{n} d_i^p \right)^{\frac{1}{p}}. \tag{17}$$

In the above equation, the exponent $p$ may further be chosen to ensure that $\mathcal{O}(\mathbf{r}_k | U_{\mathrm{ref},k}) = 0$ whenever the asset $k$ collides with any of its comparands. This condition is met if the exponent $p$ is less than or equal to zero in Eq. (17). For instance, choosing $p = 0$ yields the geometric mean (i.e., $\lim_{p \to 0} M_p = \sqrt[n]{d_1 \cdots d_n}$), while $p = -1$ gives rise to the harmonic means, just as in Eq. (13). At the limit $p \to -\infty$, Eq. (17) tends to $\min\{d_1, \ldots, d_n\}$.

Moreover, alternative formulations of the originality function can be achieved by considering different pair potentials, as noted in Sect. 4.1.

Finally, the originality functions defined in Eq. (8), (9), (11), and (13) are not bounded. An alternative definition is provided below, where the originality function $\hat{\mathcal{O}}(\mathbf{r}_k|U_{\text{ref},k})$ is bounded, taking values between 0 and 1:

$$\hat{\mathcal{O}}(\mathbf{r}_k|U_{\text{ref},k}) = \frac{\mathcal{O}(\mathbf{r}_k|U_{\text{ref},k})}{1 + \mathcal{O}(\mathbf{r}_k|U_{\text{ref},k})} = \frac{U_{\text{ref},k}}{U_{\text{ref},k} + U_k}. \tag{18}$$

However, because creativity is, in principle, not bounded, expressions as provided in Eq. (8), (9), (11), and (13), are preferred in this work.

## 5. Related work

Various originality measures have been proposed in different fields of study. To start with, originality assessments have been proposed for scientific papers, e.g., based on a structural analysis of the papers [4] or the directed citation network between references of the papers and subsequent citing papers [5]. Similarly, various indicators have been developed to measure the originality of patents [8, 9, 10]. In biology, various indices are used to measure the originality of a species, e.g., using a phylogenetic tree [6]. Creativity can also be assessed based on linguistic features of the language of study participants [11]. Originality in participant responses can further be assessed based on similarity in semantic spaces [7].

Irrespective of their merits, the schemes adopted in the above-referenced works have little in common with the present approach. Such works do not formulate the originality using concepts of maximum entropy and surprisal analysis. In particular, they do not rely on an analytical formula based on harmonic means of distances between the compared objects.

## 6. Conclusion

The present work illustrates how ML outputs and distance-computation algorithms can be exploited to assess the originality of objects such as IP assets. The originality functions are derived from concepts of maximum entropy and surprisal analysis. A simple formula is obtained, which writes as a weighted ratio of two average distances: (i) the harmonic mean of the distances between the asset of interest and its comparands (e.g., the closest prior items); and (ii) the harmonic mean of the distances between the sole comparands. Relying on harmonic mean distances ensure correct limits of the originality values. Yet, other definitions of the average can be used, which have similar properties.

The present approach allows originality scores to be assigned to the compared items. Thus, instead of comparing distances between multiple assets, originality scores can be computed and sorted, which is useful when assessing the originality of a large number of assets. This approach may potentially be used by IP professionals, IP offices, and IP courts, to gauge the originality of IP assets with respect to prior assets. More generally, the proposed schemes can be used by anyone wishing to compare originalities of items, also in a time-agnostic manner, as illustrated herein in respect of icons (smiley faces), typeface designs, paintings, and novel titles. The same approach may, in fact, be used to assess the originality of any types of digital content, including 2D or 3D images, 3D printer files, audio files, or mixed-type content. Indeed, as long as features of digital inputs can be extracted as vectors, or distances between such inputs can be computed, then their originality values can easily be computed using Eq. (11) or Eq. (13).

Of course, the proposed originality functions cannot capture all subtleties of legal analyses of IP assets. Plus, ML-based solutions remain statistical methods, which typically work well on average but are not infallible. Still, such functions allow additional facts to be obtained, which IP players may want to consider, in order to assess the value of the IP assets and the validity of the corresponding IP rights.